\documentclass[letterpaper, 10 pt, conference]{ieeeconf}  

\IEEEoverridecommandlockouts                              
\usepackage{amsmath} 
\usepackage{amssymb}  

\usepackage{amsthm}
\usepackage{algorithm}
\usepackage[noend]{algpseudocode}
\usepackage{bm}
\usepackage[pdftex]{graphicx}
\DeclareGraphicsExtensions{.pdf, .png, .jpg}
\usepackage[caption=false]{subfig}

\algrenewcommand\algorithmicindent{0.7em}%

\DeclareMathOperator*{\argmin}{arg\,\min}

\usepackage{cite}
\newtheorem{problem}{Problem}

\newtheorem{theorem}{Theorem}

\usepackage[usenames,dvipsnames]{xcolor}

\title{\LARGE \bf
Informative Planning for Worst-Case Error Minimisation in Sparse Gaussian Process Regression
}

\author{Jennifer Wakulicz, Ki Myung Brian Lee, Chanyeol Yoo, Teresa Vidal-Calleja and Robert Fitch
\thanks{This work is supported by an Australian Government Research Training Program (RTP) Scholarship and the University of Technology Sydney.}
\thanks{Authors are with the University of Technology Sydney, Ultimo, NSW 2006, Australia {\tt\footnotesize \{jennifer.wakulicz, brian.lee\}@student.uts.edu.au, \{chanyeol.yoo,teresa.vidalcalleja, rfitch\}@uts.edu.au}}%
}

\begin{document}

\maketitle
\thispagestyle{empty}
\pagestyle{empty}

\begin{abstract}
We present a planning framework for minimising the deterministic worst-case error in sparse Gaussian process~(GP) regression.
We first derive a universal worst-case error bound for sparse GP regression with bounded noise using interpolation theory on reproducing kernel Hilbert spaces~(RKHSs).
By exploiting the conditional independence~(CI) assumption central to sparse GP regression, we show that the worst-case error minimisation can be achieved by solving a posterior entropy minimisation problem.
In turn, the posterior entropy minimisation problem is solved using a Gaussian belief space planning algorithm.
We corroborate the proposed worst-case error bound in a simple 1D example, and test the planning framework in simulation for a 2D vehicle in a complex flow field.
Our results demonstrate that the proposed posterior entropy minimisation approach is effective in minimising deterministic error, and outperforms the conventional measurement entropy maximisation formulation when the inducing points are fixed. 
\end{abstract}

\section{INTRODUCTION}
Reconstructing a spatial field from sparse, noisy measurements is an important fundamental problem in robotics. The problem naturally arises in many practical applications, such as oceanography~\cite{brian2018,d2021hierarchical} and agriculture~\cite{masha_2020}, and in general tasks such as robot navigation~\cite{lan2021,maani_2017}. We are interested in informative path planning that enables robots to collect measurements for spatial field reconstruction with quality guarantees, such as minimising worst-case error. We present an approach using sparse Gaussian process~(GP) regression that is inspired by results in interpolation theory on reproducing kernel Hilbert spaces~(RKHSs). 


GP regression~\cite{rasmussen} is a powerful machine learning technique for modelling spatially correlated phenomena.
It has been widely used in the robotics community to estimate a variety of spatial fields including obstacles~\cite{lan2021,maani_2017}, infrastructure~\cite{liye_sun}, and agricultural~\cite{masha_2020} or oceanographic data~\cite{brian2018,kc_ma}.
A well-known challenge in robotics application is that the computational complexity of GP regression scales cubically with the size of the input data. 

Sparse GP approaches mitigate this computational challenge by adopting simplifying approximations~\cite{quinonero05,Wilson2015,variational,brian_icra20}.
One such approximation is the~\emph{inducing points} formulation~\cite{quinonero05}, where the target function is assumed to be conditionally independent given the function values at a fixed set of inducing points.
A recent advance in this direction is that incoming sensor measurements can be `fused' via recursive Bayesian estimation of a latent Gaussian state of \emph{fixed} dimensionality, reminiscent of Kalman filtering~\cite{recursive_sparse_gp,cadmus2021,kernel_observer,liye_sun,vidal2014,masha_2020}.

In this paper, we show that the belief maintained by a recursive Bayesian estimator is sufficient for planning paths that minimise the worst-case error in sparse GP regression with bounded noise. This result arises from interpolation theory on RKHSs, and thus assumes that the target function resides in an RKHS. This approach can be viewed as reducing the information gathering problem to a Gaussian belief space planning problem.

We present our algorithm for active spatial field reconstruction and initially demonstrate the error bound in an abstract 1D example of sparse GP regression. Then, we present results from a simulated example of path planning for an underwater robot operating in a flow field~\cite{to2019streamline,to2020distance} and collecting scalar-valued measurements. For comparison, we demonstrate the behaviour of a typical measurement entropy maximisation approach and note that the error bound is non-decreasing over time. The significance of these results is to illuminate the limitations of existing informative path planning approaches in terms of solution quality, and to contribute a new method that achieves a worst-case solution quality guarantee for actively reconstructing spatial fields.

\section{RELATED WORK}
Path planning for optimal reconstruction of a GP is typically posed as an information gain or marginal entropy maximisation problem~\cite{krause}.
These classify as a submodular maximisation problem, which is NP-hard~\cite{krause_submodular}.
When the measurement locations are subject to a dynamics constraint, a non-myopic search is necessary~\cite{graeme_acra}, which can be achieved by, e.g., growing a search tree with an appropriate pruning condition~\cite{hollinger2014,maani_2017,patten}.

Instead of the abstract information-theoretic quantities, we present an orthogonal approach that minimises the worst-case error in a \emph{deterministic} sense, inspired by the interpolation theory on RKHSs~\cite{worst_case_optimal,Karvonen2021}.
Nonetheless, we show that the worst-case error minimisation problem admits an information-theoretic analogue that is an instance of the Gaussian belief space planning~\cite{platt,atanasov2014,jen2021}.
Gaussian belief space planning is a more restricted class of problems than general submodular maximisation, and more solution algorithms are available with stronger performance guarantees.
In particular, we adopt the approach of~\cite{atanasov2014} without loss of their guarantees. 

\section{PROBLEM FORMULATION}\label{sec:prob_form}

Consider a sensor-equipped, mobile robot that operates in environment~$\mathcal{X} \subseteq \mathbb{R}^D$. The robot's motion is described by a discrete-time non-linear dynamic model
\begin{equation}\label{eq:dyn}
    \mathbf{x}_{t+1} = \mathbf{f}(\mathbf{x}_{t}, \mathbf{u}_{t})
    ,
\end{equation}
where~$\mathbf{x}_t \in \mathcal{X}$ is the state of the robot in the environment and~$\mathbf{u}_t \in \mathcal{U} \subseteq \mathbb{R}^{D_u}$ is the control input at time~$t$. A sequence of~$N$ robot states and control actions are denoted as~$\mathbf{X} = \{\mathbf{x}_0 \ldots\mathbf{x}_{N-1}\}$ and~$\mathbf{U} = \{\mathbf{u}_0 \ldots \mathbf{u}_{N-1}\}$ respectively.

The onboard sensor takes online measurements~$y_t \in \mathbb{R}$ of a scalar spatial phenomenon $s(\mathbf{x}_t)$ with additive noise~$\epsilon_t$ according to the following measurement model:
\begin{equation}\label{eq:meas}
    y_{t} = s(\mathbf{x}_{t}) + \epsilon_{t}
    .
\end{equation}

Given measurement set corresponding to a sequence of robot states $\mathbf{X}$, denoted as~$\mathbf{y}_{\mathbf{X}}$, the estimate of spatial phenomenon~$s$ over~$\mathcal{X}$ is denoted as~$\hat{s}(\mathbf{x} \mid \mathbf{y}_{\mathbf{X}})$. 
The deterministic error between true spatial phenomenon~$s$ and estimate~$\hat{s}$ is the point-wise absolute difference, defined as
\begin{equation}\label{eq:epsilon_error}
    E( \mathbf{x} \mid \mathbf{y}_{\mathbf{X}} ) = |s(\mathbf{x}) - \mathbb{E}[\hat{s}(\mathbf{x} \mid \mathbf{y}_{\mathbf{X}})]|
    .
\end{equation}

The objective of this paper is to find a sequence of control actions~$\mathbf{U}$ over time horizon~$N$ that minimises overall deterministic error~$E$ over entire domain~$\mathcal{X}$. The formal problem statement is found below.
\begin{problem}\label{prob:error}
    Given the dynamic model in~\eqref{eq:dyn} and the measurement model in~\eqref{eq:meas}, find a sequence of control actions $\mathbf{U}^*$ that minimises the total deterministic error~\eqref{eq:epsilon_error} after time-step $N$ over the domain~$\mathcal{X}$:
    \begin{equation}\begin{aligned}
        \min_{\mathbf{U}\in\mathcal{U}^{N}} \quad & \int_{\mathcal{X}} E(\mathbf{x} \mid \mathbf{y}_{\mathbf{X}}) d\mathbf{x} \\
    \end{aligned}\end{equation}
\end{problem}

Difficulty arises in solving Problem ~\ref{prob:error} in practice, as full knowledge of the ground truth $s(\mathbf{x})$ is naturally unavailable. 
Deterministic error and thus the integral in Problem ~\ref{prob:error} cannot be evaluated directly in such cases.

\section{WORST-CASE ERROR MINIMISATION AND INFORMATION GATHERING}\label{sec:reduction}

The aforementioned difficulty in solving Problem~\ref{prob:error} can be side-stepped by bounding deterministic error~\eqref{eq:epsilon_error} with an expression independent of ground truth knowledge. 
We use the tools from interpolation theory on RKHSs to find such bounds, in turn reducing Problem~\ref{prob:error} to a new entropy-based minimisation problem that is tractable.

 

\subsection{Sparse GP Regression}\label{sec:reduction:sparse_gp}
Given a sequence of noisy measurements $\mathbf{y}_{\mathbf{X}}$, we generate an estimate $\hat{s}(\mathbf{x} \mid \mathbf{y}_{\mathbf{X}}$) of the spatial phenomenon~$s(\mathbf{x})$ using GP regression with sparse approximation.
A GP is a generalisation of multivariate Gaussian random variables~(RVs) to \emph{random functions}.
A GP $s(\mathbf{x}) \sim GP(m(\mathbf{x}), k(\mathbf{x}, \mathbf{x}'))$ is characterised by a mean function $m(\mathbf{x})$ and a \emph{covariance} function $k(\mathbf{x}, \mathbf{x}')$, which specifies the covariance between function values at different points $\mathbf{x}$ and $\mathbf{x}'$~\cite{rasmussen}:
\begin{equation}\label{eq:gp_definition}
    \begin{array}{ll}
        \mathbb{E}[s(\mathbf{x})] = m(\mathbf{x}), &
        \textrm{Cov}[s(\mathbf{x}), s(\mathbf{x}')] = k(\mathbf{x}, \mathbf{x}').
    \end{array}
\end{equation}

We impose a zero-mean GP prior on the scalar field of interest, $s(\mathbf{x}) \sim GP(0, k(\mathbf{x}, \mathbf{x}'))$, with~$k(\mathbf{x}, \mathbf{x}')$ specified by the user.
Let $\mathbf{y}_{\mathbf{X}}$ be a vector containing noisy measurements up to time-step $N$ as per measurement model~\eqref{eq:meas}, i.e. $[\mathbf{y}_{\mathbf{X}}]_{i} = y_{i}$. 
With the zero-mean prior, the estimate $\hat{s}(\mathbf{x} \mid \mathbf{y}_{\mathbf{X}})$ given the measurements $\mathbf{y}_{\mathbf{X}}$ is given by another GP~\cite{kanagawa2018}:
\begin{equation}\label{eq:full_gp}
    \begin{aligned}
        \hat{s}(\mathbf{x} \mid \mathbf{y}_{\mathbf{X}}) &\sim GP( \mu (\mathbf{x} \mid \mathbf{y}_{\mathbf{X}}), \sigma^{2}(\mathbf{x}, \mathbf{x}' \mid \mathbf{y}_{\mathbf{X}})), \\
        \mu(\mathbf{x} \mid \mathbf{y}_{\mathbf{X}}) &= \mathbf{k}_{\mathbf{X}}^{\mathrm{T}}(\mathbf{x}) K_{\mathbf{X}}^{-1} \mathbf{y}_{\mathbf{X}}, \\
        \sigma^{2}(\mathbf{x}, \mathbf{x}' \mid \mathbf{y}_{\mathbf{X}}) &= k(\mathbf{x}, \mathbf{x}') -  \mathbf{k}_{\mathbf{X}}^{\mathrm{T}}(\mathbf{x}) K_{\mathbf{X}}^{-1} \mathbf{k}_{\mathbf{X}}(\mathbf{x}'),
    \end{aligned}
\end{equation}
where $[ \mathbf{k}_{\mathbf{X}}(\mathbf{x}) ]_{i} = k(\mathbf{x}, \mathbf{x}_{i})$, and $[K_{\mathbf{X}}]_{i,j} = k(\mathbf{x}_{i}, \mathbf{x}_{j})$. 
We use the same notation for other sets throughout the paper. 

Inversion of the matrix $K_{\mathbf{X}}$ in~\eqref{eq:full_gp} is computationally taxing at $\mathcal{O}(N^3)$. We use the \emph{inducing point}-based approximation of the regression~\eqref{eq:full_gp} introduced in~\cite{quinonero05} for its reduced computational complexity.
Intuitively, this formulation introduces a small set of inducing points $\mathbf{Z}=\{\mathbf{z}_{i}\}_{i=1}^{M}$ ($M \ll N$) whose function values $[\mathbf{y}_{\mathbf{Z}}]_{i} = s(\mathbf{z}_{i})$ `summarise' the entire set of measurements $\mathbf{y}_{\mathbf{X}}$. One way to assert this is to impose that $\mathbf{y}_{\mathbf{X}}$ are \emph{conditionally independent} (CI) given $\mathbf{y}_{\mathbf{Z}}$. 
Mathematically, the CI property holds if and only if:
\begin{equation}\label{eq:cond_ind}
    k(\mathbf{x}, \mathbf{x}') = \mathbf{k}_{\mathbf{Z}}^{T}(\mathbf{x}) K_{\mathbf{Z}}^{-1} \mathbf{k}_{\mathbf{Z}}(\mathbf{x}') \quad \forall \mathbf{x} \neq \mathbf{x}',
\end{equation}
which follows from asserting that the conditional cross-covariance vanishes given inducing measurements $\mathbf{y}_{\mathbf{Z}}$, i.e. $\sigma^{2}(\mathbf{x}, \mathbf{x}' \mid \mathbf{y}_{\mathbf{Z}}) = 0$. 
In other words, the correlation between any two measurements is indirect and is limited by their correlation to the inducing measurements,~$\mathbf{y}_{\mathbf{Z}}$.

We consider two popular inducing point-based approximations that satisfy CI: the subset of regressors~(SoR) and fully independent conditional~(FIC) approximations.
As noted in~\cite{quinonero05}, the SoR and FIC approximations are equivalent to replacing the kernel $k(\mathbf{x}, \mathbf{x}')$ with approximate ones as follows:
\fontsize{9}{10}
\begin{equation}\label{eq:approx_kernels}
\begin{aligned}
    \hat{k}_{SoR}(\mathbf{x}, \mathbf{x}') &= \mathbf{k}_{\mathbf{Z}}^{\mathrm{T}}(\mathbf{x})K_{\mathbf{Z}}^{-1}\mathbf{k}_{\mathbf{Z}}(\mathbf{x}'), \\
    \hat{k}_{FIC}(\mathbf{x}, \mathbf{x}') &= \hat{k}_{SoR}(\mathbf{x}, \mathbf{x}') + \delta(\mathbf{x},\mathbf{x}')(k(\mathbf{x}, \mathbf{x}') - \hat{k}_{SoR}(\mathbf{x}, \mathbf{x}')),
\end{aligned}
\end{equation}
\fontsize{10}{10}
where $\delta(\cdot)$ is the Kronecker delta function. Then, complexity reduces to $\mathcal{O}(M^2N)$ where $M \ll N$, a significant reduction compared to full GP regression~\eqref{eq:full_gp}.  

\subsection{Worst-case Error Bounds}
We use the tools from interpolation theory on RKHSs~\cite{Karvonen2021} to derive worst-case bounds on the deterministic error~\eqref{eq:epsilon_error} of sparse GP regression.
The connection between GPs and RKHSs derive from an alternate interpretation of the covariance function in a GP~\eqref{eq:gp_definition} as a \emph{positive-definite kernel}\footnote{A function $k: \mathcal{X} \times \mathcal{X} \rightarrow \mathbb{R}$ is positive definite if, for any choice of $\mathbf{X} \subset \mathcal{X}$, the matrix $[K_{\mathbf{X}}]_{i,j} = k(\mathbf{x}_{i}, \mathbf{x}_{j})$ is positive definite.}.
Any positive-definite kernel $k$ uniquely defines an RKHS $\mathcal{H}_{k}$. 
An RKHS $\mathcal{H}_{k}$ is a space of real-valued functions equipped with an inner product $\langle\cdot, \cdot\rangle_{\mathcal{H}_{k}}$ such that: 1) $k(\cdot,\mathbf{x}) \in \mathcal{H}_k$ $\forall \mathbf{x} \in \mathcal{X}$, and 2) $\langle f, k(\cdot, \mathbf{x}) \rangle_{\mathcal{H}_{k}} = f(\mathbf{x})$ $\forall \mathbf{x} \in \mathcal{X}$, $\forall \mathbf{f} \in \mathcal{H}_{k}$.
That is, the kernel function $k(\cdot, \mathbf{x})$ is itself an element of $\mathcal{H}_{k}$, and `reproduces' all other functions in $\mathcal{H}_k$ at $\mathbf{x}$.
The inner product induces a norm $||f||_{\mathcal{H}_k} = \sqrt{\langle f, f\rangle_{\mathcal{H}_{k}}}$.
In the absence of noise, it can be shown that the predictive mean of GP regression~\eqref{eq:full_gp} is exactly the minimum-norm interpolant, i.e., the function in the RKHS $\mathcal{H}_{k}$ that agrees with all measurements $\mathbf{y}_{\mathbf{X}}$ and has the minimum norm~\cite{kanagawa2018,wahba1990}.
This RKHS view of GPs is beneficial because its deterministic nature allows bounding the deterministic error~\eqref{eq:epsilon_error}, as was done in~\cite{Karvonen2021} for interpolation on RKHSs.
Inspired by such work, we establish the following worst-case error bound for GP regression with measurements containing bounded noise, which is representative of robotics applications:
\begin{theorem}\label{thm:kernel_error_bound}
Suppose $s \in \mathcal{H}_{k}$ with arbitrary positive definite kernel $k$. With bounded measurement noise $\epsilon^{2} < \sigma^{2}_{\epsilon}$, 
\begin{equation}\label{eq:error_bound_full}
    E(\mathbf{x} \mid \mathbf{y}_{\mathbf{X}})
    \leq 
    ||s||_{\mathcal{H}_{k}}
    P_{\mathbf{X}}(\mathbf{x}) 
    + \sqrt{\sigma^2_{\epsilon} N \Lambda^2_{k}(\mathbf{x})},
\end{equation}
where $P_{\mathbf{X}}(\mathbf{x}) = \sqrt{\sigma^{2}(\mathbf{x},\mathbf{x} \mid \mathbf{y}_{\mathbf{X}} )}$ is called the power function of $\mathbf{X}$ and $\Lambda_{k}(\mathbf{x}) = ||K_{\mathbf{X}}^{-1}\mathbf{k}_{\mathbf{X}}(\mathbf{x})|| $.
\end{theorem}

From Theorem~\ref{thm:kernel_error_bound}, it is clear that Problem~\ref{prob:error} can be solved by choosing a set of measurement points $\{ \mathbf{x}_{t} \}$ that minimise $P_{\mathbf{X}}(\mathbf{x})$ for all possible $\mathbf{x}$. 
A common approach in interpolation theory is to use the relationship~\cite{Karvonen2021}:
\begin{equation}\label{eq:power_function}
    P_{\mathbf{X}}(\mathbf{x}) = \sqrt{ \frac{\det K_{\mathbf{X} \cup \{\mathbf{x}\}}}{\det K_{\mathbf{X}}}},
\end{equation}
and reduce $P_{\mathbf{X}}(\mathbf{x})$ by maximising the denominator $\det K_{\mathbf{X}}$, which is independent of query point $\mathbf{x}$. 
Because Theorem~\ref{thm:kernel_error_bound} holds for an arbitrary kernel $k$, the same approach holds true for the approximate kernels~\eqref{eq:approx_kernels}.
In fact, maximising $\det K_{\mathbf{X}}$ is equivalent to measurement entropy maximisation from the informative path planning literature, e.g.,~\cite{krause}.  
However, this is still an unsatisfying answer, because 1) the numerator $\det K_{\mathbf{X} \cup \{\mathbf{x}\}}$ still varies with the choice of measurements, and 2) the choice of inducing points also affects $K_{\mathbf{X}}$.

To mitigate this issue, we exploit the CI property of sparse approximations~\eqref{eq:cond_ind}.
Because CI kernels can be viewed as interpolants to the true kernel~\cite{Wilson2015}, the interpolation of $s(\mathbf{x})$ given $\mathbf{y}_{\mathbf{X}}$ can be decomposed into two stages:
1) the interpolation of inducing measurements $\mathbf{y}_{\mathbf{Z}}$ given $\mathbf{y}_{\mathbf{X}}$, and 2) the interpolation of $s(\mathbf{x})$ given $\mathbf{y}_{\mathbf{Z}}$.
Then, it is natural to ask if the deterministic error~\eqref{eq:error_bound_full} or the power function~\eqref{eq:power_function} admits a similar decomposition.
The following theorem confirms that there is such a decomposition. 

\begin{theorem}\label{thm:sparse_error_bound}
Suppose a kernel $k$ satisfies the CI assumption~\eqref{eq:cond_ind}. Then, the power function $P_{\mathbf{X}}(\mathbf{x})$ satisfies:
\begin{equation}\label{eq:PF_result}
    \begin{aligned}
        P_{\mathbf{Z}}(\mathbf{x}) \leq P_{\mathbf{X}}(\mathbf{x}) \leq P_{\mathbf{Z}}(\mathbf{x}) \exp H(\mathbf{y}_{\mathbf{Z}} \mid \mathbf{y}_{\mathbf{X}}),
    \end{aligned}
\end{equation}  
where $P_{\mathbf{Z}}(\mathbf{x})$ is the power function of $\mathbf{Z}$ as per~\eqref{eq:power_function}
and $H(\mathbf{y}_{\mathbf{Z}}\mid\mathbf{y}_{\mathbf{X}})$ is the posterior entropy of $\mathbf{y}_{\mathbf{Z}}$ given $\mathbf{y}_{\mathbf{X}}$:
\begin{equation}\label{eq:posterior_entropy}
    H(\mathbf{y}_{\mathbf{Z}}\mid\mathbf{y}_{\mathbf{X}}) = \frac{1}{2} \log\left( (2\pi e)^{M} \frac{\det K_{\mathbf{Z} \cup \mathbf{X}} }{\det K_{\mathbf{X}}} \right).
\end{equation}

Moreover, assuming $s \in \mathcal{H}_{k}$, the deterministic error~\eqref{eq:epsilon_error} can be further bounded as:
\begin{equation}\label{eq:error_bound_ci}
    E(\mathbf{x}|\mathbf{y}_{\mathbf{X}}) \leq 
    ||s||_{\mathcal{H}_{k}} 
    P_{\mathbf{Z}}(\mathbf{x}) \exp H(\mathbf{y}_{\mathbf{Z}} \mid \mathbf{y}_{\mathbf{X}})
    + \sqrt{\sigma_{\epsilon}^2 N \Lambda^2_{k}(\mathbf{x})}
\end{equation}
\end{theorem}

Using Theorem~\ref{thm:sparse_error_bound}, we approximately solve Problem~\ref{prob:error} via the following surrogate problem, which minimises the control-dependent terms in the new upper bound~\eqref{eq:error_bound_ci}: 
\begin{problem} \label{prob:det}
Given the dynamic model~\eqref{eq:dyn} and the measurement model~\eqref{eq:meas}, find a sequence of control actions $\mathbf{U}^{*}$ that minimises the posterior entropy of inducing measurements:
\begin{equation}\label{eq:posterior_entropy_minimise}
    \min_{\mathbf{U}\in \mathcal{U}^N} H(\mathbf{y}_{\mathbf{Z}} \mid \mathbf{y}_{\mathbf{X}}).
\end{equation}
\end{problem}

The merit of the reformulation in Problem~\ref{prob:det} is that the worst-case total deterministic error~\eqref{eq:error_bound_full} can be approximately minimised without explicitly integrating over the operating region in the noise-free case.
This is because the posterior entropy $H(\mathbf{y}_{\mathbf{Z}} \mid \mathbf{y}_{\mathbf{X}})$ is the only term dependent on the control actions, but is independent of the integrand $\mathbf{x}$.
Further, Problem~\ref{prob:det} serves as a closer proxy to Problem~\ref{prob:error} as $H(\mathbf{y}_{\mathbf{Z}} \mid \mathbf{y}_{\mathbf{X}}) \rightarrow 0$ (i.e., as more measurements are added, or with better solution quality).
This is owing to the `tightness' of the inequality~\eqref{eq:PF_result} in that $P_{\mathbf{X}}(\mathbf{x}) \rightarrow P_{\mathbf{Z}}(\mathbf{x})$ as~$H(\mathbf{y}_{\mathbf{Z}} \mid \mathbf{y}_{\mathbf{X}}) \rightarrow 0$.
The fact that $P_{\mathbf{X}}(\mathbf{x})$ approaches $P_{\mathbf{Z}}(\mathbf{x})$ illustrates the importance of selecting good inducing points $\mathbf{Z}$ with low $P_{\mathbf{Z}}(\mathbf{x})$.




\section{PLANNING FRAMEWORK}\label{sec:algorithm}
A further benefit of the reformulation in Problem~\ref{prob:det} is that we need only to compute and minimise the posterior entropy of a fixed number of variables (i.e., the inducing measurements $\mathbf{y}_{\mathbf{Z}}$), unlike previous formulations~\cite{krause,kc_ma} where the dimensionality grows.  
We exploit this benefit by using an efficient recursive sparse GP algorithm presented in~\cite{recursive_sparse_gp} that maintains a Gaussian belief over the inducing measurements $\mathbf{y}_{\mathbf{Z}}$.
In turn, the Gaussian representation allows the use of the reduced value iteration~(RVI) algorithm~\cite{atanasov2014} for minimising the posterior entropy while retaining its strong guarantees.

\subsection{Recursive Sparse GP Regression}
Given measurements up to time $t$, the recursive sparse GP regression algorithm~\cite{recursive_sparse_gp} permits equivalent calculation to GP regression~\eqref{eq:full_gp}, while only storing the posterior mean and covariance of the inducing measurements $\mathbf{y}_{\mathbf{Z}}$:
\begin{equation}
\begin{array}{ll}
    \bm{\mu}_{t} = \mathbb{E}[\mathbf{y}_{\mathbf{Z}} \mid \mathbf{y}_{t}], & \Sigma_{t} = \textrm{Cov}[\mathbf{y}_{\mathbf{Z}} \mid \mathbf{y}_{t}].   
\end{array}
\end{equation}
Importantly, the posterior entropy can be calculated as a function of $\Sigma_{t}$:
\begin{equation}\label{eq:posterior_entropy_recursive}
H(\mathbf{y}_{\mathbf{Z}} \mid \mathbf{y}_{{\mathbf{X}}}) \equiv c(\Sigma_{t}) = \frac{1}{2} \log \det 2 \pi e \Sigma_{t}. 
\end{equation}
This is more efficient than direct computation using~\eqref{eq:posterior_entropy}, and has additional benefit of fitting directly into the sparse recursive GP regression algorithm as follows.

Letting $\mathbf{q}(\mathbf{x}) = K_{\mathbf{Z}}^{-1}\mathbf{k}_{\mathbf{Z}}(\mathbf{x})$, we can recover the posterior GP~\eqref{eq:full_gp} (i.e., perform regression) given the belief $\bm{\mu}_{t}$ and $\Sigma_{t}$:
\begin{equation}\label{eq:sparse_gp_predict}
\begin{aligned}
    \mu(\mathbf{x} \mid \mathbf{y}_{t}) &= \mathbf{q}^{T}(\mathbf{x}) \bm{\mu}, \\
    \sigma^{2}_{SoR}(\mathbf{x}, \mathbf{x}' \mid \mathbf{y}_{t}) &= \mathbf{q}^{T}(\mathbf{x})\Sigma_{t}\mathbf{q}(\mathbf{x}'), \\
    \sigma^{2}_{FIC}(\mathbf{x}, \mathbf{x}' \mid \mathbf{y}_{t}) &= \sigma^{2}_{SoR}(\mathbf{x} \mathbf{x}' \mid \mathbf{y}_{t}), \\
                                                                &+\delta(\mathbf{x}, \mathbf{x}')(k(\mathbf{x}, \mathbf{x}') - \hat{k}_{SoR}(\mathbf{x}, \mathbf{x}')).
\end{aligned}
\end{equation}



The update procedure is analogous to a Kalman filter.
Initially, the belief is set to $\bm{\mu}_{0}=0$ and $\Sigma_{0} = K_{\mathbf{Z}}$. 
Given measurement~$y_{t}$ at $\mathbf{x}_{t}$, we generate the predictive mean, variance and cross-covariance at future time-step:
\begin{equation}\label{eq:kf_prediction}
\begin{aligned}
    \hat{y}_{t+1} &= \mu(\mathbf{x}_{t} \mid \mathbf{y}_{t}), \\
    \Sigma^{yy}_{t+1} &= \sigma^{2}_{*}(\mathbf{x}_{t}, \mathbf{x}_{t}) + \sigma_{\epsilon}^{2}, \\
    \Sigma^{y\mathbf{Z}}_{t+1} &= \mathbf{q}(\mathbf{x}_{t}) \Sigma_{t}.
\end{aligned}
\end{equation} 
      Using the predictions~\eqref{eq:kf_prediction}, and new measurement~$y_{t+1}$ at~$\mathbf{x}_{t+1}$ we perform a Kalman update:
\begin{equation}\label{eq:kalman}
\begin{aligned}
    \bm{\mu}_{t+1} &= \bm{\mu}_{t} + \Sigma^{y\mathbf{Z}}_{t+1} (\Sigma^{yy}_{t+1})^{-1} (y_{t+1} - \hat{y}_{t+1}), \\ 
    \Sigma_{t+1} &= \Sigma_{t} - \Sigma^{y\mathbf{Z}}_{t+1} (\Sigma^{yy}_{t+1})^{-1} (\Sigma^{y\mathbf{Z}}_{t+1})^{T}.
\end{aligned}
\end{equation}
As with the Kalman filter, here the covariance prediction and update~\eqref{eq:kf_prediction}-\eqref{eq:kalman} are measurement independent, enabling posterior entropy~\eqref{eq:posterior_entropy_recursive} minimisation over non-myopic horizons.

\subsection{Receding Horizon Planning}
Given the belief maintained by recursive GP, we plan a path that minimises the posterior entropy~\eqref{eq:posterior_entropy_recursive} using an adapted version of the RVI algorithm~\cite{atanasov2014}.
The RVI algorithm maintains a search tree $\mathcal{T}$ over possible trajectories and belief. 
Each node $v \in \mathcal{T}$ is associated with a candidate robot position and predicted posterior covariance $(\mathbf{x}_{t}^{v}, \Sigma_{t}^{v})$. 

Alg.~\ref{alg:overview} shows an operational overview. 
The tree is initialised with the robot's initial position and belief $(\mathbf{x}_{0}, \Sigma_{0})$~(line~\ref{alg:overview:init}).
Initially, we perform an offline search over horizon $N$ by repeatedly iterating the RVI algorithm~(Alg.~\ref{alg:rvi}) in line~\ref{alg:overview:offline}.
Each RVI iteration~(Alg.~\ref{alg:rvi}) expands the search tree by one timestep and adds the corresponding layer of leaves.

During the online stage, the robot extracts the optimal control from the tree by searching for the lowest-cost leaf node (lines~\ref{alg:overview:extract_1}-\ref{alg:overview:extract_2}).
The corresponding control action $\mathbf{u}^{*}$ is executed, and the robot reaches a new state $\mathbf{x}_{t+1}$ and obtains a new measurement $y_{t+1}$~(line~\ref{alg:overview:execute}).
The belief $(\bm{\mu}_{t+1}, \Sigma_{t+1})$ is updated using the recursive GP equation~\eqref{eq:kalman}~(line~\ref{alg:overview:update}).
To generate a new plan, we re-use the subtree $\mathcal{T}_{l^{*}}$ rooted at the chosen node $l^{*}$ and perform the RVI iteration~(line~\ref{alg:iterate}).
Because the RVI iteration adds a new layer of leaves, the depth of the tree is always equal to time horizon $N$.

The RVI iteration proceeds as follows.
First, a new layer of leaves are expanded from the current leaves $\mathcal{L}(\mathcal{T})$ in lines~\ref{alg:rvi:for_leaves}-~\ref{alg:rvi:add_to_tree} by sampling the control space $\mathcal{U}$ and propagating the state $\mathbf{x}_{N-1}$ and posterior covariance $\mathbf{\Sigma}_{N-1}$ forward. 
Note that the propagation of posterior covariance $\mathbf{\Sigma}_{N-1}$ only depends on $\mathbf{x}_{N-1}$, and does not require the measurement $y_{N-1}$. 

Next, we iterate over the newly added leaves and extract the set of nodes $Q$ whose state are within $\delta$-distance of each other (line~\ref{alg:rvi:delta}).
If such nodes exist, we check for $\epsilon$-algebraic redundancy~($\epsilon$-alg. red.) (line~\ref{alg:rvi:epsilon}). 
A node $l$ is $\epsilon$-alg. red. iff there exists a set of coefficients $\{\alpha_{q}\}_{i=1}^{|Q|}$ such that $\sum \alpha_{q} = 1$ and $\Sigma_T + \epsilon I \succeq \sum_{i=1}^{|Q|}\alpha_i\Sigma_i$. 
A candidate node is pruned if it is within $\delta$-distance of other leaf nodes, and is also $\epsilon$-alg. red. with respect to those nodes (line~\ref{alg:rvi:prune}).

The benefit of RVI is the strong suboptimality bound that accompanies it. 
The cost $c(\Sigma_{N}^{RVI})$ returned by RVI and the optimal cost $c(\Sigma_{N}^{*})$ satisfy $0 < c(\Sigma^{*}_{N}) - c(\Sigma^{RVI}_{N}) < C(\epsilon, \delta)$, where $C$ is a problem-specific function.
In particular, with $\epsilon,\delta=0$ the result is optimal~\cite{atanasov2014}.
Note that the same bound holds for Problem~\ref{prob:det}, because it is of the same form as in~\cite{atanasov2014}.

As noted in~\cite{atanasov_thesis}, the $\epsilon$-alg. red. check is an instance of LMI feasibility problem, and poses computational challenge as the number of inducing points grows.
The challenge can be circumvented by setting $\epsilon = \infty$.
In this case, the RVI iteration~(Alg.~\ref{alg:rvi}) only adds the lowest cost nodes that are not within $\delta$ distance of each other, owing to the ascending order of iteration~(line~\ref{alg:rvi:ascending}).
While there are no bounds in this case, it produces practically viable solutions.

\begin{algorithm}[t]
\caption{Receding horizon planning for worst-case error minimisation}\label{alg:overview}
\begin{algorithmic}[1]
\State $\mathcal{T} \gets \{(\mathbf{x}_{0}, \Sigma_{0})\}$\label{alg:overview:init}
\For{$t = 1, ..., N$}\label{alg:overview:offline}
    \State $\mathcal{T} \gets \texttt{RVI}(\mathcal{T})$
\EndFor
\While{robot is operational}
    \State $l^{*} \gets \argmin_{l \in \mathcal{L}(\mathcal{T})} c(\Sigma^{l}) $\label{alg:overview:extract_1}
    \State $\mathbf{u}^{*} \gets \texttt{backtrace}(l^{*}) $\label{alg:overview:extract_2}
    \State $\mathbf{x}_{t+1}, y_{t+1} \gets$ execute $\mathbf{u}^{*}$ and sample measurement\label{alg:overview:execute}
    \State Update $\bm{\mu}_{t+1}, \Sigma_{t+1}$ with $y_{t+1}$ using \eqref{eq:kalman}\label{alg:overview:update}
    \State $\mathcal{T} \gets \texttt{RVI}(\mathcal{T}_{l^{*}})$\label{alg:iterate}
\EndWhile
\end{algorithmic}
\end{algorithm}
\begin{algorithm}[t]
    \caption{Reduced value iteration}\label{alg:rvi}
    \begin{algorithmic}[1]
        \For{$\forall l \in \mathcal{L}(\mathcal{T})$}\label{alg:rvi:for_leaves}
            \For{$\forall \mathbf{u}_{N-1} \in \mathcal{U}$}
                \State $\mathbf{x}^{u}_{N} \gets \mathbf{f}(\mathbf{x}_{N-1}^{l}, \mathbf{u}_{N-1})$
                \State $\Sigma^{u}_{N} \gets$ update $\Sigma^{l}_{N-1}$ with~\eqref{eq:kalman}
                \State $\mathcal{N} \gets \mathcal{T} \cup \{ (\mathbf{x}^{u}_{N}, \Sigma^{u}_{N}) \}$\label{alg:rvi:add_to_tree}
            \EndFor
        \EndFor
        \State $S_{\min} \gets \{ l \in \mathcal{L}(\mathcal{T}) \mid \Sigma^{l}_{N} = \argmin c(\Sigma_{N})\}$
        \For{$l \in \mathcal{L}(\mathcal{T}) \setminus S_{\min}$ in ascending order of $c(\Sigma_{N})$} \label{alg:rvi:ascending}
            \State $Q \gets \{\Sigma_{N}^{u} \mid u \in \mathcal{L}(\mathcal{T}), d(\mathbf{x}^{l}_{N}, \mathbf{x}^{u}_{N}) \leq \delta \}$\label{alg:rvi:delta}
            \If $Q$ is not empty and $\Sigma_{N}$ is $\epsilon$-alg. red. w.r.t. $Q$\label{alg:rvi:epsilon}
                \State $\mathcal{T} \gets \mathcal{T} \setminus l$\label{alg:rvi:prune}
            \EndIf
        \EndFor
        \State Return $\mathcal{T}$
    \end{algorithmic}
\end{algorithm}




\begin{figure}[!t]
    \centering
    \includegraphics{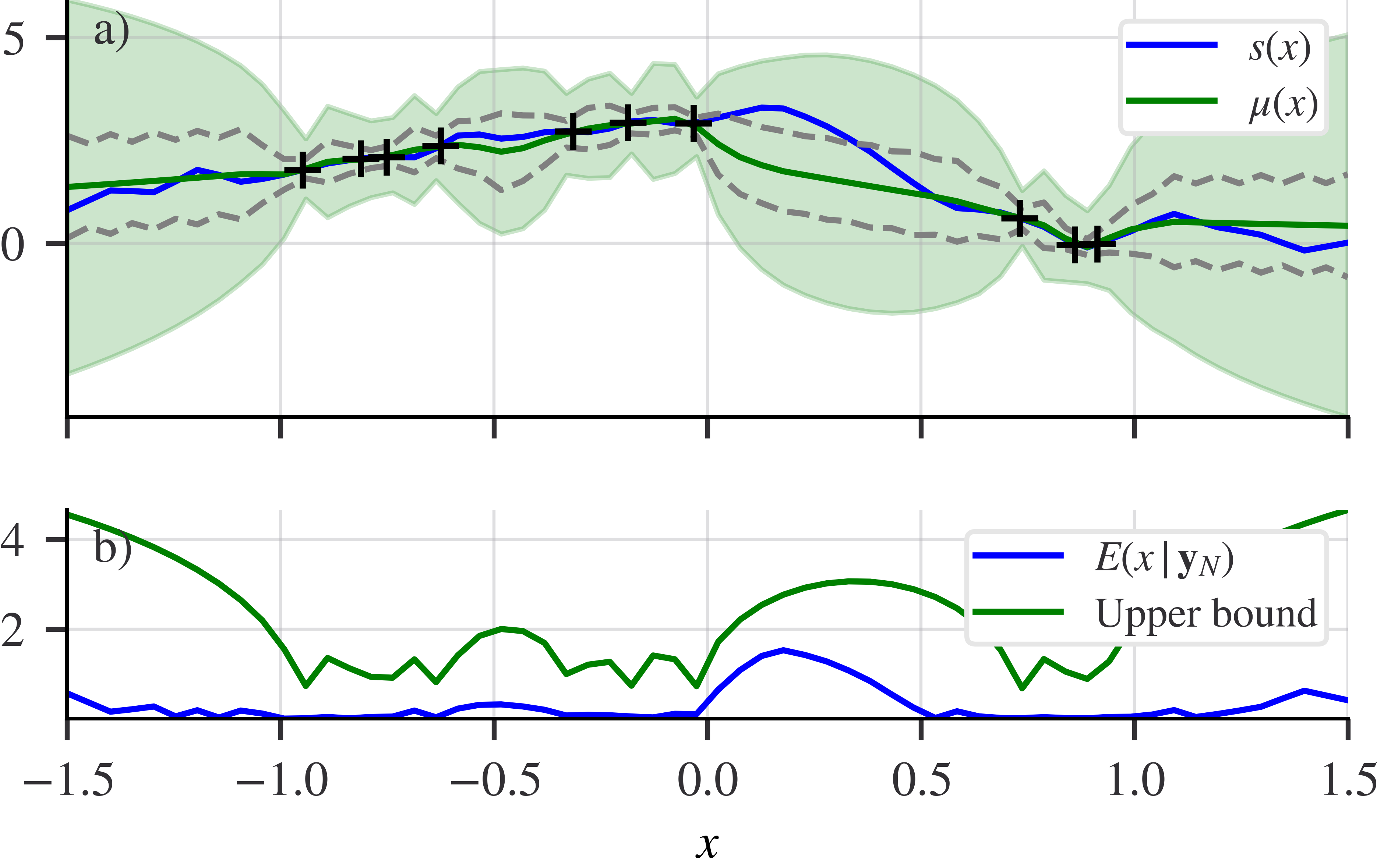}
    \vspace{-1ex}
    \caption{a) One-dimensional sparse Gaussian process regression of target function in the RKHS. Measurement locations are indicated with black crosses. The proposed bound on deterministic error is shown in the green shaded area. The area enclosed by the grey dotted lines is the standard $1\sigma$-confidence interval. b) Demonstration of the upper bound on deterministic error point-wise over the domain.}\label{fig:bound_demo}
    \vspace{-3ex}
\end{figure}

\section{EXPERIMENTAL RESULTS}
\subsection{Characterisation of the Error Bound}
We first corroborate the error bound proposed in Theorem~\ref{thm:kernel_error_bound} with an example. 
As the RKHS norm required for computing the error bound may be hard to compute, particularly in higher dimensional settings, we consider a one-dimensional regression problem where the target function is designed with simplified form $s(x) = \sum_{i=1}^m \alpha_i k(x,x_i). $
This way we ensure $s\in\mathcal{H}_K$, and the RKHS norm is easily reduced to the Euclidean norm. Such simplifications are enough to demonstrate the tightness and behaviour of the proposed bound. 

Fig.~\ref{fig:bound_demo}a) depicts the outcome of regression with such a target function. With sparse and noisy measurements, the GP regressor is able to reconstruct the target function well. Importantly, the proposed bounds on deterministic error are reasonably tight on the predicted mean and follow the intuitive behaviour of decreasing near measurement locations. When compared to the $1\sigma$-confidence interval obtained from the posterior covariance, the proposed bounds have overall higher value. However, in certain regions of sparse or no measurement the target function is greater than the $1\sigma$-confidence interval and yet remains within our error bound. This demonstrates that while confidence intervals may be broken, the error bound may not.

In Fig.~\ref{fig:bound_demo}b), we show the deterministic error~\eqref{eq:epsilon_error} against the proposed upper bound, confirming the deterministic error lies below the bound for all $x$ in the domain. This result corroborates that our bound does indeed give a reasonably tight upper bound on deterministic error. 

\subsection{Flow Field Case Study}

\begin{figure}
    \centering
    \includegraphics[width=0.6\columnwidth]{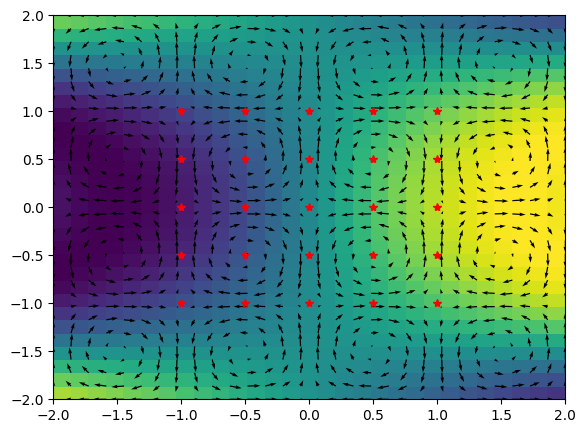}
    \caption{The ground truth environment: brighter colour indicates higher value, inducing points used for sparse GP regression are shown in red.}
    \label{fig:ground_truth}
    \vspace{-3ex}
\end{figure}

To demonstrate capability of the proposed algorithm and problem formulation in information-theoretic path planning, we consider a simplified underwater glider operating in a double-gyre flow field.
The dynamics are given by:
\begin{equation}
\begin{bmatrix}
    x_{t+1} \\
    y_{t+1}
\end{bmatrix}
    = \begin{bmatrix}
    x_{t} \\
    y_{t}
\end{bmatrix} + \Delta t (V_{g} \begin{bmatrix}
    -\sin(\pi x) \cos(\pi y) \\
    \cos(\pi x) \sin(\pi y)
\end{bmatrix}  + V \begin{bmatrix}
    \cos{u_{t}} \\
    sin{u_{t}} 
\end{bmatrix}).
\end{equation}
The robot aims to solve Problem~\ref{prob:det} equipped with the planning algorithm given in Alg.~\ref{alg:overview} in order to reconstruct a scalar field of interest (such as level of salinity) over the flow field. The scalar field is shown in Fig.~\ref{fig:ground_truth} as a colour map superimposed on the flow field.

\begin{figure}[h!]
    \centering
    \includegraphics[width=\columnwidth]{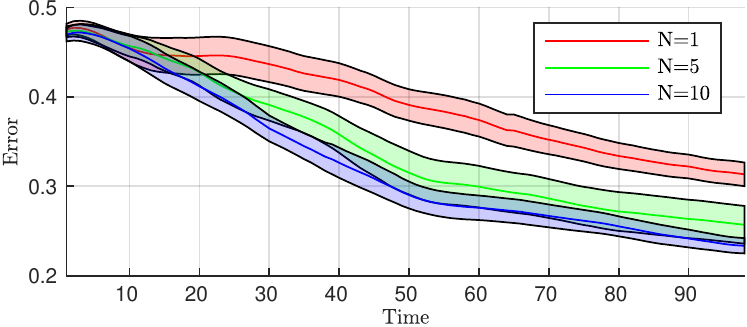}
    \vspace{-4ex}
    \caption{Average absolute reconstruction error with varying search horizon. Shaded areas represent 95\% confidence interval over 20 trials.}
    \label{fig:logdet:error}
    \vspace{-3ex}
\end{figure}

To verify that the algorithm solves the deterministic error minimisation problem (Problem~\ref{prob:error}), we evaluate the average absolute error sampled over a 30$\times$30 grid. 
We vary the horizon $N$ between $\{1, 5, 10\}$ and examine the average absolute error over time. 
Fig.~\ref{fig:logdet:error} shows the result over 20 randomised initial starting locations in the same environment~(Fig.~\ref{fig:ground_truth}).
It can be seen that for all choices of search horizon, the average absolute error decreases over time. 
The rate of reduction is greater with larger search horizon. 

\begin{figure}[t!]
    \centering
        \subfloat[$N=1$]{\includegraphics[width=0.49\columnwidth]{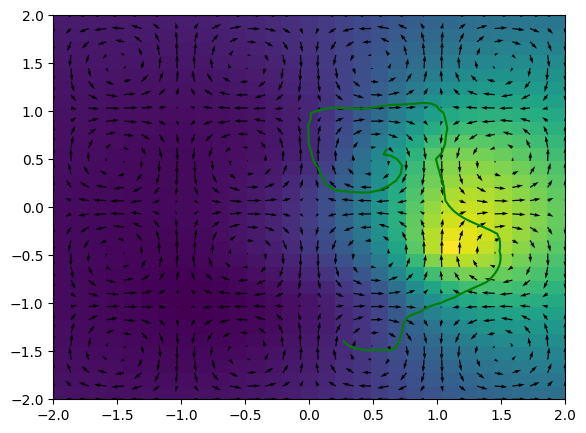}\label{fig:logdet:h1:m}}    
        \subfloat[$N=1$]{\includegraphics[width=0.49\columnwidth]{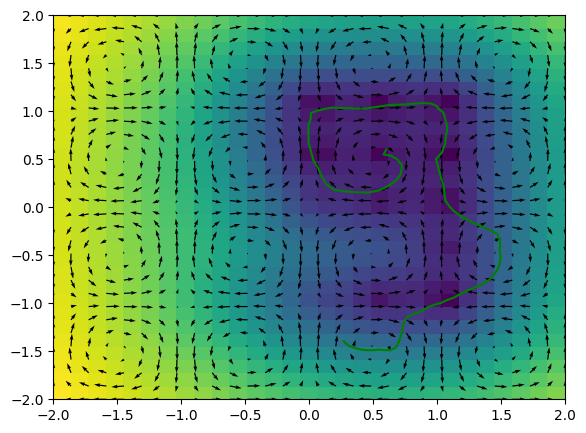}\label{fig:logdet:h1:v}} \\
        \vspace{-2ex}
        \subfloat[$N=5$]{\includegraphics[width=0.49\columnwidth]{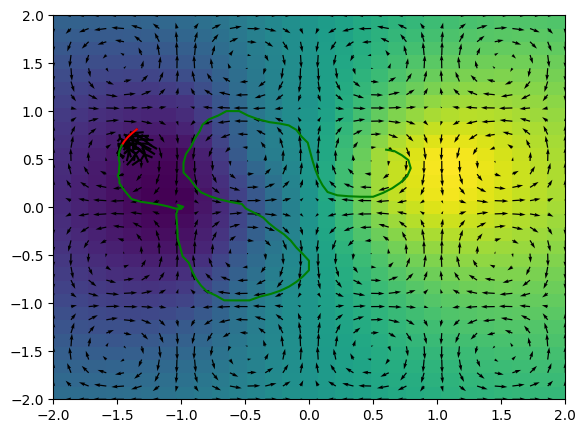}\label{fig:logdet:h5:m}} 
        \subfloat[$N=5$]{\includegraphics[width=0.49\columnwidth]{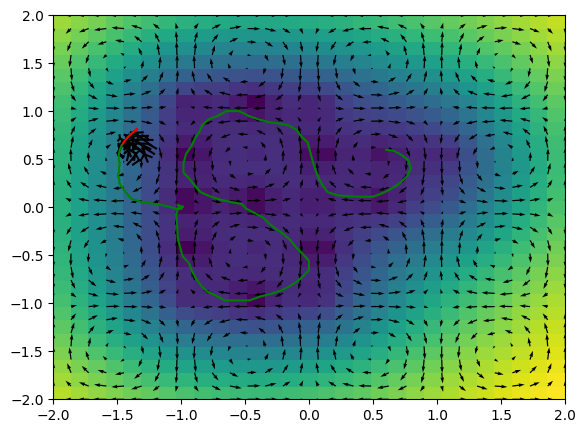}\label{fig:logdet:h5:v}} \\
        \vspace{-2ex}
        \subfloat[$N=10$]{\includegraphics[width=0.49\columnwidth]{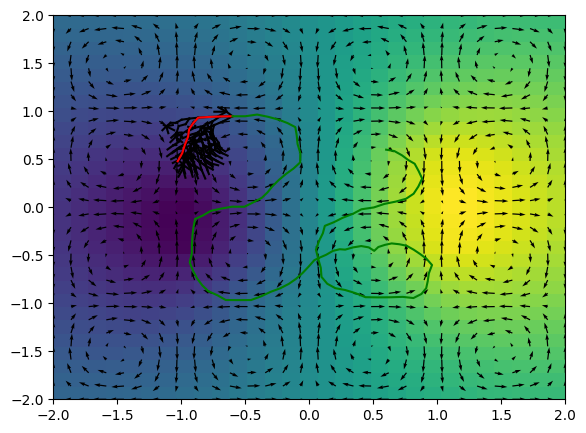}\label{fig:logdet:h10:m}} 
        \subfloat[$N=10$]{\includegraphics[width=0.49\columnwidth]{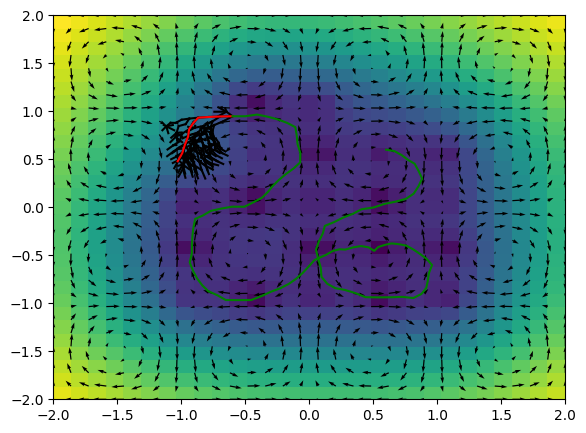}\label{fig:logdet:h10:v}} 
    \caption{Example trajectories after 100 steps with varying search horizon. Green: executed trajectory. Left column: mean ($\mu(\mathbf{x} \mid \mathbf{y}_{\mathbf{X}})$). Right column: variance, or square of power function ($\sigma^{2}(\mathbf{x} \mid \mathbf{y}_{\mathbf{X}}) = P_{\mathbf{X}}^{2}(\mathbf{x})$). Red: current plan. Black: search tree.  Brighter colour means higher value. Non-myopy leads to better coverage and reconstruction of the function of interest.}
    \label{fig:logdet}
    \vspace{-1ex}
\end{figure}
\begin{figure}
    \centering
    \includegraphics[width=\columnwidth]{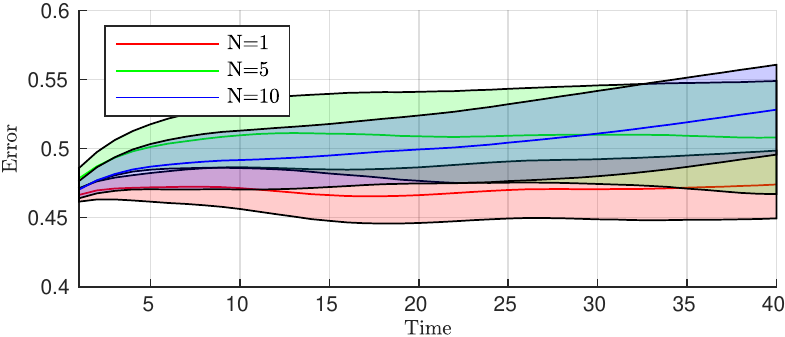}
    \vspace{-4ex}
    \caption{Average absolute reconstruction error with varying search horizon for the measurement entropy maximisation formulation. Shaded areas represent 95\% confidence interval over 20 trials.}
    \vspace{-3ex}
    \label{fig:maxdet:error}
\end{figure}
\begin{figure}
    \centering
    \subfloat[]{\includegraphics[width=0.24\textwidth]{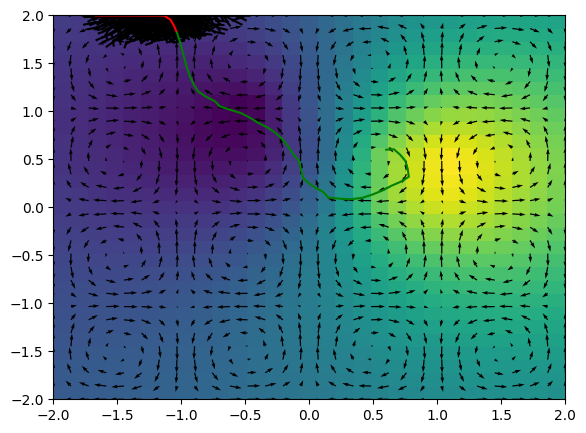}\label{fig:maxdet:h5:m}}
    \subfloat[]{\includegraphics[width=0.24\textwidth]{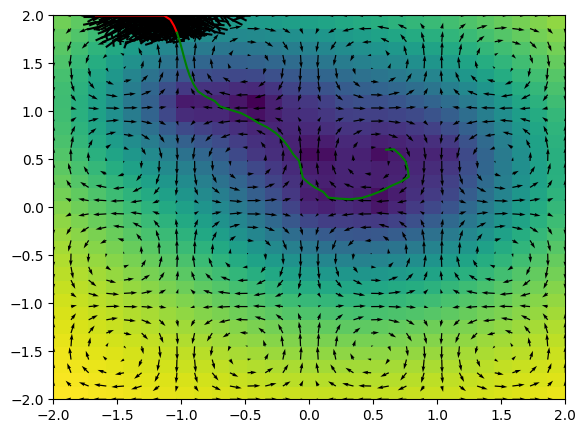}\label{fig:maxdet:h5:v}} 
    \caption{An example trajectory from measurement entropy maximisation. (a): mean ($\mu(\mathbf{x} \mid \mathbf{y}_{\mathbf{X}})$). (b): variance, or square of power function ($\sigma^{2}(\mathbf{x} \mid \mathbf{y}_{\mathbf{X}}) = P_{\mathbf{X}}^{2}(\mathbf{x})$) Green: executed trajectory. Red: current plan. Black: search tree. Measurement entropy maximisation compels the robot to always explore outward. However, further away from the inducing points, the variance never decrease regardless of measurements.}
    \label{fig:maxdet}
    \vspace{-3ex}
\end{figure}

To better understand this behaviour, we show example trajectories after 100 time steps in Fig.~\ref{fig:logdet}. 
For the myopic greedy horizon case in Fig.~\ref{fig:logdet:h1:m},~\ref{fig:logdet:h1:v} we observe poor coverage of the spatial domain. Additionally, the myopic nature of this planner results in trajectories that often get ``stuck" in an attracting region of a gyre. The robot is unable to use the flow field dynamics to its advantage to best explore, and reconstruction of the spatial field is thus poor. 

With longer planning horizons the robot successfully manoeuvres through the flow field to increase coverage of the domain, giving improved estimation of the spatial field as seen in Fig.~\ref{fig:logdet:h5:m} and~\ref{fig:logdet:h10:m}, with best estimation and coverage given by the longest horizon $N=10$. Fig.~\ref{fig:logdet} further demonstrates the influence of the objective~\eqref{prob:det} on trajectories. For all horizon lengths, the robot preferentially takes measurements near inducing points over exploring regions further away, such that variance is minimised at inducing points. This is exemplified in Fig.~\ref{fig:logdet} where with increasing horizon, broader coverage of the region around inducing points and greater minimisation of variance is achieved.

\subsection{Comparison to Measurement Entropy Maximisation}
Under the same experimental scenario~(Fig.~\ref{fig:ground_truth}), we examine the behaviour of the solution of entropy maximisation, which is the usual approach to information gathering in GPs~\cite{krause}.
This was implemented by setting the cost in Alg.~\ref{alg:rvi} as $-\log \det K_{\mathbf{X}}$ with $\epsilon = \infty$. 

Fig.~\ref{fig:maxdet:error} shows the average absolute error over the environment with 20 random initial positions and varying search horizon.
Surprisingly, the error does not decrease over time, and actually increases with larger search horizons~($N=10$).

To understand the finding, we examine an example trajectory after $40$ time-steps in Fig.~\ref{fig:maxdet:error}, where the robot expands outwards making use of the ambient flow field.
This is expected because the measurement entropy maximisation formulation demands the robot to simply move as far away from its previous trajectory as possible. 
However, the expansive behaviour is problematic when using inducing point-based GPs.
As the robot gets further from the inducing points, the measurements do not make a significant contribution.
Estimates further away from inducing points remain to be of poor quality regardless of measurements taken, as seen near the end of the green trajectory in Fig.~\ref{fig:maxdet:h5:v}. 
Coupled with measurement entropy maximisation, the robot expands outwards in a positive feedback loop. 
A possible solution is to re-adjust the inducing points in an online manner via gradient updates~\cite{kc_ma}.
We defer online update of inducing points to future work.

\section{CONCLUSION}
We derived an upper bound to worst-case deterministic error for sparse GP regression with bounded noise. 
We proved this upper bound naturally gives way to an information-theoretic analogue to minimisation of deterministic error. 
Thus, one may minimise deterministic error via an information-theoretic proxy. 
Our results demonstrate the proposed approach outperforms conventional methods in reducing deterministic error. 
Results illuminated clear limitations involving placement of sparse GP inducing points which will be addressed in future work via online updates to inducing point locations. 
Further work also lies in extensions for reconstruction of time-varying  and multi-dimensional spatial fields~\cite{brian2019,cadmus2021,yoo2019stochastic,yoo2021path,lee2020hierarchical}.

\section*{APPENDIX}
\begin{proof}[Proof of Theorem 1] Letting $[\mathbf{s}(\mathbf{x}_N)]_i = s(\mathbf{x}_i)$, $i = 1,\ldots,N$ be the vector of function evaluations at $\mathbf{x}_N$, and similarly $[\bm{\epsilon}_N]_i = \epsilon_i$, we have:
    \begin{equation*}
    \begin{aligned}
        &E(\mathbf{x} \mid \mathbf{y}_{\mathbf{X}}) = |s(\mathbf{x}) - \mu(\mathbf{x} \mid \mathbf{y}_{\mathbf{X}})|\\
        &\leq |s(\mathbf{x}) - \mathbf{k}_{\mathbf{X}}^{\mathrm{T}}(\mathbf{x})K^{-1}_{\mathbf{X}}\mathbf{s}(\mathbf{x}_N)| +\\ 
        &\hspace{7mm}|\mathbf{k}_{\mathbf{X}}^{\mathrm{T}}(\mathbf{x})K^{-1}_{\mathbf{X}}\mathbf{s}(\mathbf{x}_N) - \mathbf{k}_{\mathbf{X}}^{\mathrm{T}}(\mathbf{x})K^{-1}_{\mathbf{X}}[\mathbf{s}(\mathbf{x}_N) + \bm{\epsilon}_N]|\\
        &= |\langle s, k(\cdot,\mathbf{x}) - \mathbf{k}_{\mathbf{X}}^{\mathrm{T}}(\mathbf{x})K^{-1}_{\mathbf{X}}\mathbf{k}_\mathbf{X}(\cdot)\rangle_{\mathcal{H}_k}|+|\mathbf{k}_{\mathbf{X}}^{\mathrm{T}}(\mathbf{x})K^{-1}_{\mathbf{X}}\bm{\epsilon}_N|\\
        &\leq ||s||_{\mathcal{H}_k}P_{\mathbf{X}}(\mathbf{x}) +|\mathbf{k}_{\mathbf{X}}^{\mathrm{T}}(\mathbf{x})K^{-1}_{\mathbf{X}}\bm{\epsilon}_N|\\
        &\leq ||s||_{\mathcal{H}_k}P_{\mathbf{X}}(\mathbf{x}) + 
        \sqrt{\sigma_{\epsilon}^2 N}||\mathbf{k}_{\mathbf{X}}^{\mathrm{T}}(\mathbf{x})K^{-1}_{\mathbf{X}}||
    \end{aligned}
    \end{equation*}
where the final inequality follows from $|\cdot| = \sqrt{\langle\cdot,\cdot\rangle}$, Cauchy-Schwarz inequality and the assumption $\epsilon_i^2 < \sigma_{\epsilon}^2$ $\forall i = 1,\ldots,N$.
\end{proof}

\begin{proof}[Proof of Theorem 2] We exploit the fact that the power function $P_N(\mathbf{x})$ can be linked to conditional entropy as:
\begin{equation*}
    H(s(\mathbf{x})|\mathbf{y}_{\mathbf{X}}) = \frac{1}{2}(\log P_{\mathbf{X}}(\mathbf{x}) + \log(2\pi e)).
\end{equation*}
Using the CI assumption, it can be shown that:
\begin{equation}
    H(s(\mathbf{x})|\mathbf{y}_{\mathbf{X}}) = H(s(\mathbf{x})|\mathbf{y}_{\mathbf{Z}}) + \mathcal{I}(\mathbf{y}_{\mathbf{Z}};s(\mathbf{x})|\mathbf{y}_{\mathbf{X}}).
\end{equation}
It then follows that:
\begin{equation}\label{eq:proof_eq}
    \begin{aligned}
    H(s(\mathbf{x})|\mathbf{y}_{\mathbf{Z}}) \leq H(s(\mathbf{x})|\mathbf{y}_{\mathbf{X}}) \leq H(s(\mathbf{x})|\mathbf{y}_{\mathbf{Z}}) + H(\mathbf{y}_{\mathbf{Z}}|\mathbf{y}_{\mathbf{X}}),
    \end{aligned}
\end{equation}
and the claimed bound is recovered by taking the exponential of~\eqref{eq:proof_eq}.
\end{proof}

\bibliography{reference}

\begin{thebibliography}{10}
\providecommand{\url}[1]{#1}
\csname url@samestyle\endcsname
\providecommand{\newblock}{\relax}
\providecommand{\bibinfo}[2]{#2}
\providecommand{\BIBentrySTDinterwordspacing}{\spaceskip=0pt\relax}
\providecommand{\BIBentryALTinterwordstretchfactor}{4}
\providecommand{\BIBentryALTinterwordspacing}{\spaceskip=\fontdimen2\font plus
\BIBentryALTinterwordstretchfactor\fontdimen3\font minus
  \fontdimen4\font\relax}
\providecommand{\BIBforeignlanguage}[2]{{%
\expandafter\ifx\csname l@#1\endcsname\relax
\typeout{** WARNING: IEEEtran.bst: No hyphenation pattern has been}%
\typeout{** loaded for the language `#1'. Using the pattern for}%
\typeout{** the default language instead.}%
\else
\language=\csname l@#1\endcsname
\fi
#2}}
\providecommand{\BIBdecl}{\relax}
\BIBdecl

\bibitem{brian2018}
K.~M.~B. Lee, J.~J.~H. Lee, C.~Yoo, B.~Hollings, and R.~Fitch, ``Active
  perception for plume source localisation with underwater gliders,'' in
  \emph{Proc. of ARAA ACRA}, 2018.

\bibitem{d2021hierarchical}
G.~D’urso, J.~J.~H. Lee, O.~Pizarro, C.~Yoo, and R.~Fitch, ``Hierarchical
  mcts for scalable multi-vessel multi-float systems,'' in \emph{Proc. of IEEE
  ICRA}.\hskip 1em plus 0.5em minus 0.4em\relax IEEE, 2021, pp. 8664--8670.

\bibitem{masha_2020}
M.~Popovic, T.~Vidal-Calleja, G.~Hiltz, J.~J. Chung, I.~Sa, R.~Siegwart, and
  J.~Nieto, ``An informative path planning framework for {UAV}-based terrain
  monitoring,'' \emph{Auton. Rob.}, vol.~44, pp. 889 -- 911, 2020.

\bibitem{lan2021}
L.~Wu, K.~M.~B. Lee, L.~Liu, and T.~Vidal-Calleja, ``Faithful {E}uclidean
  distance field from log-{G}aussian process implicit surfaces,'' \emph{Rob.
  and Auotom. Lett.}, vol.~6, pp. 2461--2468, 2021.

\bibitem{maani_2017}
M.~G. Jadidi, J.~V. Miro, and G.~Dissanayake, ``Gaussian processes autonomous
  mapping and exploration for range-sensing mobile robots,'' \emph{Auton.
  Rob.}, vol.~42, pp. 273--290, 2020.

\bibitem{rasmussen}
C.~K. Williams and C.~E. Rasmussen, \emph{Gaussian processes for machine
  learning}.\hskip 1em plus 0.5em minus 0.4em\relax MIT press Cambridge, MA,
  2006, vol.~2, no.~3.

\bibitem{liye_sun}
L.~Sun, T.~Vidal-Calleja, and J.~V. Miro, ``Bayesian fusion using conditionally
  independent submaps for high resolution {2.5D} mapping,'' in \emph{Proc. of
  IEEE ICRA}, 2015.

\bibitem{kc_ma}
K.-C. Ma, L.~Liu, H.~K. Heidarsson, and G.~S. Sukhatme, ``Data-driven learning
  and planning for environmental sampling,'' \emph{J. Field Robot.}, vol.~35,
  pp. 643--661, 2018.

\bibitem{quinonero05}
J.~Quinonero-Candela and C.~E. Rasumussen, ``A unifying view of sparse
  approximate {G}aussian process regression,'' \emph{J. Mach. Learn. Res.},
  vol.~6, pp. 1939--1959, 2005.

\bibitem{Wilson2015}
A.~G. Wilson and H.~Nickisch, ``{Kernel interpolation for scalable structured
  Gaussian processes (KISS-GP)},'' \emph{32nd International Conference on
  Machine Learning, ICML 2015}, vol.~3, pp. 1775--1784, 2015.

\bibitem{variational}
M.~Titsias, ``Variational learning of inducing variables in sparse {G}aussian
  processes,'' in \emph{Proc. of Artificial intelligence and statistics}.\hskip
  1em plus 0.5em minus 0.4em\relax PMLR, 2009, pp. 567--574.

\bibitem{brian_icra20}
K.~M.~B. Lee, W.~Martens, J.~Khatkar, R.~Fitch, and R.~Mettu, ``Efficient
  updates for data association with mixtures of {Gaussian} processes.''\hskip
  1em plus 0.5em minus 0.4em\relax IEEE, 2020, pp. 335--341.

\bibitem{recursive_sparse_gp}
M.~Schürch, D.~Azzimonti, A.~Benavoli, and M.~Zaffalon, ``Recursive estimation
  for sparse {Gaussian} process regression,'' \emph{Automatica}, vol. 120, p.
  109127, 10 2020.

\bibitem{cadmus2021}
K.~Y.~C. To, F.~H. Kong, K.~M.~B. Lee, C.~Yoo, S.~Anstee, and R.~Fitch,
  ``Estimation of spatially correlated ocean currents from ensemble forecasts
  and online measurements,'' in \emph{Proc. of IEEE ICRA}, 2021.

\bibitem{kernel_observer}
H.~A. Kingravi, H.~Maske, and G.~Chowdhary, ``{Kernel observers:
  Systems-theoretic modeling and inference of spatiotemporally evolving
  processes},'' in \emph{Proc. of Adv. Neural Inf. Process. Syst.}\hskip 1em
  plus 0.5em minus 0.4em\relax Curran Associates, Inc., 2016, pp. 3990--3998.

\bibitem{vidal2014}
T.~Vidal-Calleja, D.~Su, F.~De~Bruijn, and J.~V. Miro, ``{Learning spatial
  correlations for {Bayesian} fusion in pipe thickness mapping},'' in
  \emph{Proc. of IEEE ICRA}.\hskip 1em plus 0.5em minus 0.4em\relax IEEE, 2014,
  pp. 683--690.

\bibitem{to2019streamline}
K.~C. To, J.~J.~H. Lee, C.~Yoo, S.~Anstee, and R.~Fitch, ``Streamline-based
  control of underwater gliders in 3d environments,'' in \emph{Proc. of IEEE
  CDC}.\hskip 1em plus 0.5em minus 0.4em\relax IEEE, 2019, pp. 8303--8310.

\bibitem{to2020distance}
K.~C. To, C.~Yoo, S.~Anstee, and R.~Fitch, ``Distance and steering heuristics
  for streamline-based flow field planning,'' in \emph{Proc. of IEEE
  ICRA}.\hskip 1em plus 0.5em minus 0.4em\relax IEEE, 2020, pp. 1867--1873.

\bibitem{krause}
A.~Krause, A.~Singh, and C.~Guestrin, ``Near-optimal sensor placements in
  {Gaussian} processes: Theory, efficient algorithms and empirical studies,''
  \emph{J. Mach. Learn. Res}, vol.~9, pp. 235--284, 02 2008.

\bibitem{krause_submodular}
A.~Krause and D.~Golovin, ``Submodular function maximization.''
  \emph{Tractability}, vol.~3, pp. 71--104, 2014.

\bibitem{graeme_acra}
G.~Best and R.~Fitch, ``Probabilistic maximum set cover with path constraints
  for informative path planning,'' in \emph{Proc. of ARAA ACRA}, 2016.

\bibitem{hollinger2014}
G.~A. Hollinger and G.~S. Sukhatme, ``Sampling-based robotic information
  gathering algorithms,'' \emph{The Int. Journ. of Rob. Res.}, vol.~33, no.~9,
  pp. 1271--1287, 2014.

\bibitem{patten}
T.~Patten and R.~Fitch, ``{Monte Carlo} planning for active object
  classification,'' \emph{Auton. Robot}, vol.~42, pp. 391 -- 421, 2018.

\bibitem{worst_case_optimal}
T.~Karvonen and S.~Särkkä, ``Worst-case optimal approximation with
  increasingly flat {Gaussian} kernels,'' \emph{Adv. Comput. Math.}, vol.~46,
  03 2020.

\bibitem{Karvonen2021}
T.~Karvonen, S.~S{\"{a}}rkk{\"{a}}, and K.~Tanaka, ``{Kernel-based
  interpolation at approximate Fekete points},'' \emph{Numer. Algorithms},
  vol.~87, no.~1, pp. 445--468, 2021.

\bibitem{platt}
R.~Platt, R.~Tedrake, L.~P. Kaelbling, and T.~Lozano-Perez, ``Belief space
  planning assuming maximum likelihood,'' in \emph{Proc. of RSS}, 2010.

\bibitem{atanasov2014}
N.~Atanasov, J.~Le~Ny, K.~Daniilidis, and G.~J. Pappas, ``Information
  acquisition with sensing robots: Algorithms and error bounds,'' in
  \emph{Proc. of IEEE ICRA}, 2014, pp. 6447--6454.

\bibitem{jen2021}
J.~Wakulicz, H.~Kong, and S.~Sukkarieh, ``Active information acquisition under
  arbitrary unknown disturbances,'' in \emph{Proc. of IEEE ICRA}, 2021.

\bibitem{kanagawa2018}
M.~Kanagawa, P.~Hennig, D.~Sejdinovic, and B.~K. Sriperumbudur, ``Gaussian
  processes and kernel methods: A review on connections and equivalences,''
  2018.

\bibitem{wahba1990}
G.~Wahba, \emph{Spline models for observational data}.\hskip 1em plus 0.5em
  minus 0.4em\relax SIAM, 1990.

\bibitem{atanasov_thesis}
N.~Atanasov, ``Active information acquisition with mobile robots,'' Ph.D.
  dissertation, University of Pennsylvenia, 2015.

\bibitem{brian2019}
K.~M.~B. Lee, C.~Yoo, B.~Hollings, S.~Anstee, S.~Huang, and R.~Fitch, ``Online
  estimation of ocean current from sparse {GPS} data for underwater vehicles,''
  in \emph{Proc. of IEEE ICRA}.\hskip 1em plus 0.5em minus 0.4em\relax IEEE,
  2019, pp. 3443--3449.

\bibitem{yoo2019stochastic}
C.~Yoo, S.~Anstee, and R.~Fitch, ``Stochastic path planning for autonomous
  underwater gliders with safety constraints,'' in \emph{Proc. of IEEE/RSJ
  IROS}.\hskip 1em plus 0.5em minus 0.4em\relax IEEE, 2019, pp. 3725--3732.

\bibitem{yoo2021path}
C.~Yoo, J.~J.~H. Lee, S.~Anstee, and R.~Fitch, ``Path planning in uncertain
  ocean currents using ensemble forecasts,'' in \emph{Proc. of IEEE
  ICRA}.\hskip 1em plus 0.5em minus 0.4em\relax IEEE, 2021, pp. 8323--8329.

\bibitem{lee2020hierarchical}
J.~J.~H. Lee, C.~Yoo, S.~Anstee, and R.~Fitch, ``Hierarchical planning in
  time-dependent flow fields for marine robots,'' in \emph{Proc. of IEEE
  ICRA}.\hskip 1em plus 0.5em minus 0.4em\relax IEEE, 2020, pp. 885--891.

\end{thebibliography}

\end{document}